\documentclass{article}
\usepackage{spconf,amsmath,graphicx}
\usepackage{colortbl}
\usepackage{pifont}

\usepackage{multirow}
\usepackage{booktabs}
\usepackage[normalem]{ulem}

\title{Efficient Convolution and Transformer-Based Network for Video Frame Interpolation}
\name{Issa Khalifeh$^{1,2}$, Luka Murn$^{1}$, Marta Mrak$^{2}$, Ebroul Izquierdo$^{2}$ \thanks{This work was supported under the iCase grant (Ref: 2246814) from the Engineering and Physical Sciences Resarch Council (EPSRC) in collaboration with the British Broadcasting Cooperation (BBC)}}
\address{$^{1}$BBC Research and Development, London, United Kingdom  \\
$^{2}$Queen Mary University of London, London, United Kingdom\\}
\begin{document}
%
\maketitle

\begin{abstract}
Video frame interpolation is an increasingly important research task with several key industrial applications in the video coding, broadcast and production sectors. Recently, transformers have been introduced to the field resulting in substantial performance gains. However, this comes at a cost of greatly increased memory usage, training and inference time. In this paper, a novel method integrating a transformer encoder and convolutional features is proposed. This network reduces the memory burden by close to 50$\%$ and runs up to four times faster during inference time compared to existing transformer-based interpolation methods. A dual-encoder architecture is introduced which combines the strength of convolutions in modelling local correlations with those of the transformer for long-range dependencies. Quantitative evaluations are conducted on various benchmarks with complex motion to showcase the robustness of the proposed method, achieving competitive performance compared to state-of-the-art interpolation networks.
\end{abstract}
\begin{keywords}
Video frame interpolation, transformer, complexity reduction, dual-encoder
\end{keywords}
\section{Introduction}
\label{sec:intro}
Video frame interpolation, which aims to synthesise intermediate frames from existing ones, is a fundamental computer vision application that has seen many developments over the years. Nowadays, interpolation methods can be broadly categorised into kernel-based and flow-based methods. Many flow-based methods first estimate the optical flow and refine the flow maps using a convolutional neural network (CNN) \cite{niklaus2020softmax,kong2022ifrnet, jiang2018super}. Kernel-based methods use a U-Net architecture to extract kernels which are used as weights for a separable convolution operation \cite{cheng2021multiple,niklaus2017video, lee2020adacof}. As optical flow isn't explicitly synthesised in these techniques, issues with interpolating challenging scenes such as occlusions, brightness changes, high motion can be handled well. 

\begin{figure}
\centering
\begin{center}
\includegraphics[width=1\linewidth]{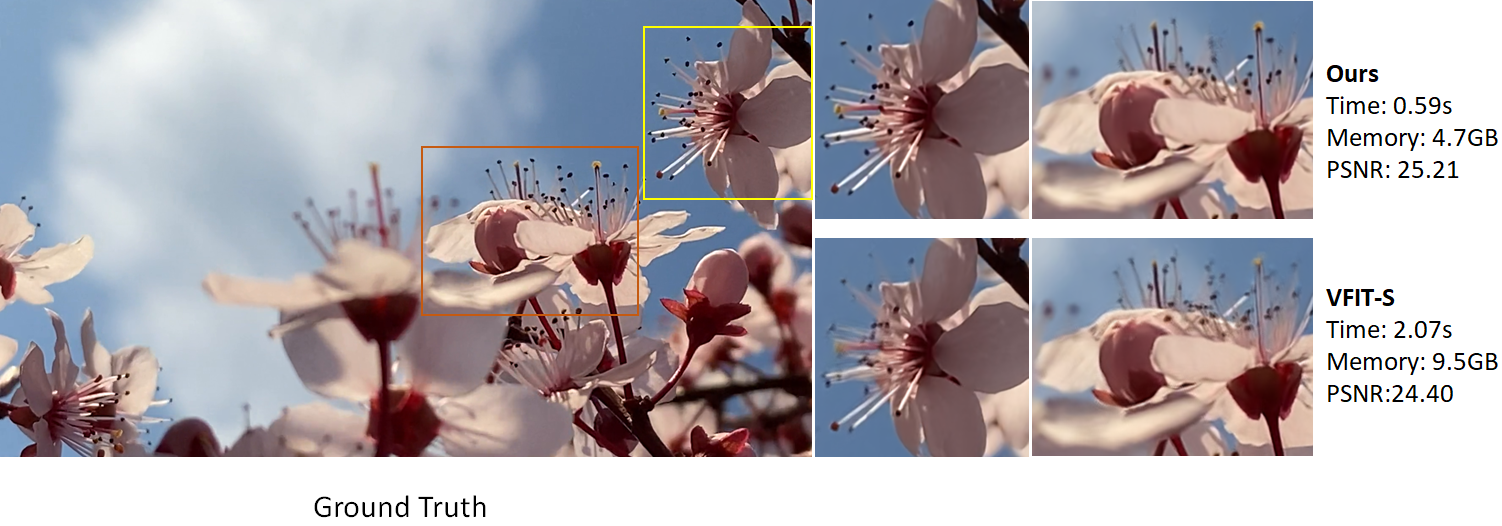}
\end{center}
\vspace{-2em}
\caption{An example showcasing a challenging sequence consisting of large pixel displacement from the VFITex test set.}
\label{fig1:comp}
\vspace{-1em}
\end{figure}
  
\begin{figure*}
\centering
\begin{center}
   \includegraphics[width=1\linewidth]{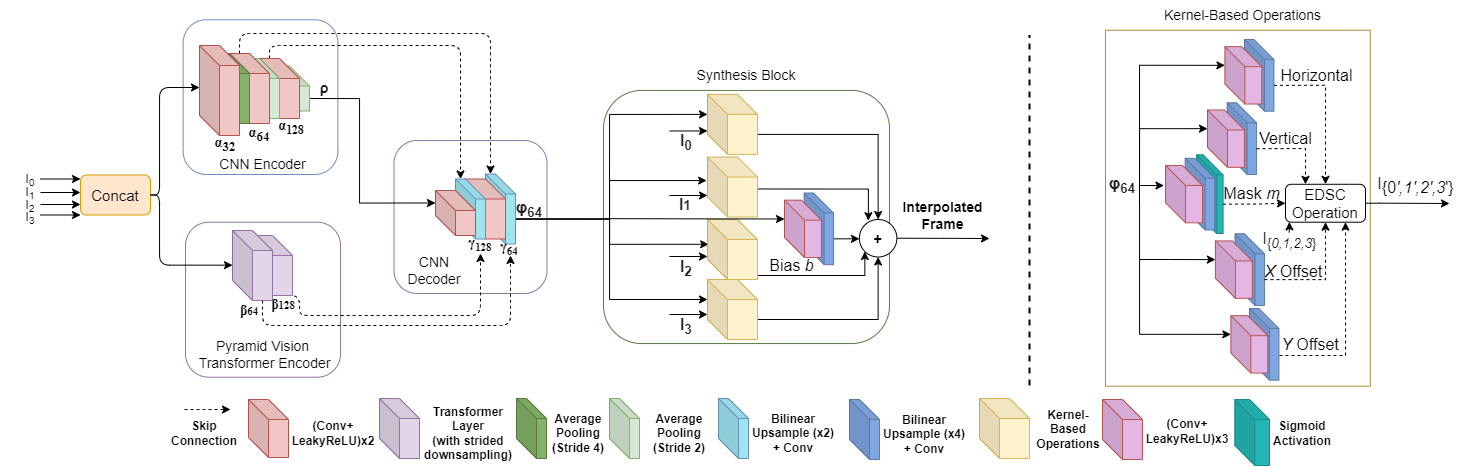}
\end{center}
\vspace{-1.5em}
\caption{Overview of the proposed network. The input frames are concatenated in the channel dimension. Concatenated frames are then used as inputs to the CNN Encoder and the Pyramid Vision Transformer Encoder. The feature maps generated at the 64 and 128 channel feature level are combined with decoder features through a skip connection. The output of the decoder is used for kernel-based operations in the Synthesis Block for the generation of the output frame.}
\label{fig2:arch}
\end{figure*}

Recently, transformers have been introduced to frame interpolation networks, namely the flow-based VFIformer \cite{lu2022video} and the kernel-based VFIT \cite{shi2022video}. VFIformer takes the generated flow and downsamples it at each stage, using it as a guiding tool for the transformer encoder. VFIT employs 3D convolutions to better model temporal correlations and modifies the Swin transformer \cite{liu2021swin} to better optimise it for the task of interpolation. VFIformer handles two input frames whereas VFIT makes use of four frames. Using four frames allows the network to better handle occlusions due to the availability of more reference points and is becoming more common in video frame interpolation models \cite{xu2019quadratic,danier2022st,kalluri2020flavr}. VFIT proposes two models, large and small. The large model has twice as many channels at each stage compared to the small model. However, memory usage and runtime are high for both versions. For the VFIT-small model, interpolating a 1080p frame on an NVIDIA Quadro RTX 5000
consumes approximately 9.5 GB of GPU memory and requires more than two seconds to synthesise an HD frame, as shown in Figure~\ref{fig1:comp}. This is not practical when interpolating long sequences and especially when using limited computational resources. 

In this work, we propose a model comparable to VFIT-small which uses 50$\%$ less memory and runs up to four times faster during inference time. This is achieved by an efficient architecture design that minimises the use of computationally intensive operations, such as 3D convolutions present throughout VFIT. We propose handling inputs using a pyramid vision transformer \cite{wang2022pvt} and adapt its architecture for the interpolation task. Convolutional and transformer features are integrated into the network design and combined to leverage the inherent advantages of these different representations. This network is termed \textbf{E}fficient \textbf{D}ual \textbf{E}ncoder \textbf{N}etwork for \textbf{V}ideo \textbf{F}rame \textbf{I}nterpolation (EDENVFI). 

\noindent The main contributions of our work are:
\begin{itemize}
\item To our knowledge, we are the first to integrate the pyramid vision transformer encoder for the task of video frame interpolation.
\item We combine transformer and convolutional features to better handle sequences with different types of motion.
\item Extensive evaluations on benchmarks demonstrate competitive performance of our fast and memory-efficient model compared to state-of-the-art methods.
\end{itemize}

\section{Proposed Approach}
\label{sec:format}
For an input set $\mathbf{I}$ of four frames ${I_0, I_1, I_2, I_3}$, an intermediate output frame $I_{out} = I_{1.5}$ needs to be synthesised. The proposed network  EDENVFI consists of four main components: the pyramid vision transformer encoder, the convolutional encoder, the convolutional decoder and the synthesis block. Figure~\ref{fig2:arch} gives an overview of the network architecture.

\textbf{Pyramid Vision Transformer (PVT) Encoder}: PVT is selected as it is flexible with different input sizes and is able to model correlations well, evidenced by its competitive performance against other methods \cite{wang2022pvt}. PVT extracts hierarchical features and models long-range dependencies which is beneficial for video frame interpolation, where both the local and global context are important for the synthesis of a more realistic frame. As PVT was designed for classification tasks, we modify the network design to be compatible with the task of interpolation. The output classes of the network are reduced from $1000$ to $1$ class. The PVT encoder is configured to have two feature levels of $64$ and $128$ channels with a layer depth of 9 and 12, respectively. Spatial reduction ratios are set to $16$ and $8$. Empirical studies conducted on deeper level networks showed an increase in runtime and memory usage for minimal performance gain.

To extract hierarchical features, strided downsampling is used in the convolution for the patch projection of transformer inputs $\mathbf{I}$. The stride in the first feature level is set to four. This significantly reduces the computational burden, especially when handling larger HD videos in the first stage of the network. As a stride of four results in the loss of information and thus a degradation in performance, the depth of the transformer is increased to ensure correlations can still be modelled adequately.

\textbf{Convolutional Encoder}: Although PVT has made great strides in modelling local as well as global features through its network design, the locality of convolutions can still provide benefits to model performance \cite{qu2022transmef}. A convolutional encoder is thus used in our network design. There are three channel levels $(32, 64, 128)$ and each convolutional block consists of two 2D convolutions followed by a leaky ReLU. After each block, average pooling is used. The stride is set to four for the first pooling operation and the stride is set to two after the output of the $64$ and $128$ channel blocks. The output from the final pooling block is termed $\mathbf{\rho}$. As with the PVT encoder, the input to the CNN encoder are the channel-wise concatenated input frames ${I_0, I_1, I_2, I_3}$.
\begin{figure*}
\centering
\begin{center}
\includegraphics[width=1\linewidth]{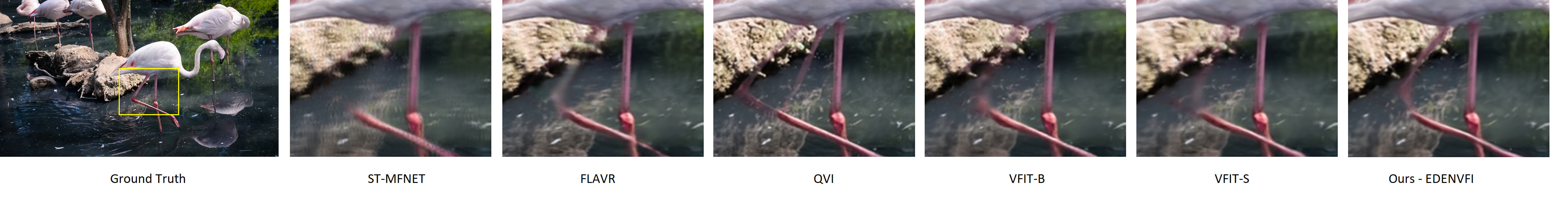}
\end{center}
\vspace{-2em}
\caption{An example showcasing a challenging sequence on the DAVIS set on state-of-the-art methods.}
\label{fig3:comp}
\vspace{-0.5em}
\end{figure*}
VFIT and other interpolation networks \cite{shi2022video,danier2022st,kalluri2020flavr} use 3D convolutions, as they are able to model the spatio-temporal relations between frames better compared to 2D convolutions. However, they are much more computationally expensive. 3x3 convolutions are used throughout our network and are supplemented by the pyramid vision transformer encoder to model long-range dependencies and improve performance.

\textbf{Convolutional Decoder}: The direct input to the decoder are the final pooled convolutional feature maps $\rho$ from the CNN encoder. The decoder consists of two convolutional blocks and two upsample blocks. The upsample block consists of a bilinear interpolation layer and a 3x3 convolution. The output from each upsample block is combined with the PVT and CNN feature maps, represented as:
\vspace{-0.5em}
\begin{equation}
\varphi_{L} = \alpha_{L} + \beta_{L} + \gamma_{L}
\vspace{-0.5em}
\label{eq1}
\end{equation}

\noindent where $\alpha_{L}$, $\beta_{L}$ $\gamma_{L}$ are convolutional, transformer and upsampled features at level $L = \{128, 64\}$. The output from the final skip connection $\varphi_{64}$ is used as an input for the synthesis block.

\textbf{Synthesis Block}: The synthesis block from \cite{cheng2021multiple} is adapted to consider four input frames. For each input frame, a set of kernels is generated using $\varphi_{64}$ features, consisting of horizontal and vertical kernels, $X$ and $Y$ offsets and a modulation mask. The generated features are then used as weights for a separable deformable convolution operation of kernel size $5$ which is applied to each patch of the input image. After applying the deformable convolution on ${I_0, I_1, I_2, I_3}$, the output is combined and a bias is added to obtain the final output. An example operation of the synthesis block is shown in Figure~\ref{fig2:arch}.

The enhanced deformable separable convolution operation (EDSC) can be represented as follows:
\vspace{-0.5em}
\begin{equation}
I_{out} = \sum_{{i=0}}^{3} \mathbf{K}_{i}(x,y) \cdot \mathbf{K}_i^{T}(x,y) \ast \mathbf{P'}_{i}(x,y) + b
\vspace{-0.5em}
\label{equationkernel}
\end{equation}

\noindent where $I_{out}$ is the output interpolated frame, $\mathbf{K}$ and $\mathbf{K}^{T}$ are the horizontal and vertical kernels, $x$ and $y$ are coordinates in the output frame, $\mathbf{P'}$ is the image patch where convolution will occur, and $b$ is the bias which is the image residual directly generated by the network. Horizontal and vertical offsets $X$ and $Y$ are applied to local patches $P$ on positions $x$ and $y$ to obtain $\mathbf{P'}$ which is defined as:
\vspace{-0.5em}
\begin{equation}
\mathbf{P'}(x,y) = P(x,y: x+X,y+Y) \cdot m
\vspace{-0.5em}
\label{equationpatch}
\end{equation}
\noindent where $m$ is the modulation mask.

\section{Experiments}
\vspace{-0.5em}
\subsection{Training Procedure and Evaluation Sets}
\vspace{-0.5em}
EDENVFI is trained for 100 epochs using the AdaMAX optimiser \cite{kingma2014adam}. Finding optimal hyper-parameters for interpolation networks is challenging. VFIT uses a tailored learning rate configuration where the learning rate is set to specific values upon reaching milestone epochs. For our proposed network, an initial rate of 0.0005 was found to yield better performance compared to the 0.001 learning rate, typically used for training video frame interpolation networks \cite{cheng2021multiple, lee2020adacof}. The learning rate halves when performance on the validation set does not increase after 5 epochs. L1 loss is used as $L_{1} = \| I_{out} - I_{gt}\|$, where $I_{gt}$ is the ground truth.

The Vimeo90K septuplet \cite{xue2019video} set is used for training and evaluation. UCF-101 and DAVIS selections from \cite{xu2019quadratic} are used as evaluation sets. As noted in \cite{danier2022st}, Vimeo90K does not contain much dynamic motion, so the SNU-FILM set \cite{choi2020channel}, divided into easy, medium, hard and extreme parts, and the VFITex dataset also used for evaluation.
\vspace{-0.5em}
\subsection{Ablation Study}
\vspace{-0.5em}
\label{sec:pagestyle}

Table~\ref{tab1} shows model performance for different transformer layer depth configurations. Network performance generally increases with the increase in depth  up to $(15,15)$. The $(9,12)$ configuration is chosen as our preferred one to balance runtime and performance requirements. 

The effectiveness of the convolutional and PVT encoder features is investigated in Table~\ref{tab3}. Integrating PVT yields performance gains across all the datasets compared to the convolution-only model. This gain is most notable with Davis where results are more than 1 dB higher. The PVT-only model has a slightly higher performance on the SNU-FILM set relative to the combined attention model. However, the performance of the combined attention model on Vimeo90K and DAVIS exceeds that of the transformer-only model. This indicates that introducing a convolutional encoder may have a more noticeable performance gain when handling constrained motions similar to those in Vimeo90K as opposed to more dynamic motions as in SNU-FILM.

\begin{table}
\vspace{-1.5em}
\centering
\caption{The impact of increasing the pyramid vision transformer layer depth on PSNR performance on UCF101, DAVIS and Vimeo90k datasets.
}
\resizebox{\columnwidth}{!}{
\begin{tabular}{ccclc} 
\midrule
Parameters (M) & \begin{tabular}[c]{@{}c@{}}PVT Layer\\Depth\end{tabular} & UCF101 & \multicolumn{1}{c}{DAVIS} & Vimeo90K  \\ 
\midrule
2.21           & 0                                                        & 33.025 & 26.917                    & 35.416    \\
10.96           & 3,4                                                      & 33.174 & 27.655                    & 35.901    \\
27.59           & 9,12                                                     & 33.311 & 28.130                    & 36.332    \\
30.89           & 12,12                                                    & 33.277 & 28.045                    & 36.301    \\
37.95           & 15,15                                                    & 33.354 & 28.158                    & 36.387    \\
\midrule
\end{tabular}}
\label{tab1}
\vspace{-1.5em}
\end{table}

\begin{table*}
\centering
\caption{Impact of the convolutional and transformer encoders on network performance on various testsets}
\begin{tabular}{cc|cccccccl} 
\midrule
\multicolumn{2}{c}{Encoder}                        & \multicolumn{4}{c}{SNU-FILM}       & \multirow{2}{*}{UCF101} & \multirow{2}{*}{DAVIS} & \multirow{2}{*}{Vimeo90K} & \multicolumn{1}{c}{\multirow{2}{*}{VFITex}}  \\ 
\cmidrule{1-6}
\multicolumn{1}{l}{Conv} & \multicolumn{1}{l}{PVT} & Extreme & Hard   & Medium & Easy   &                         &                        &                           & \multicolumn{1}{c}{}                         \\ 
\midrule
\ding{51}                       & ~                       & 24.843  & 30.563 & 36.339 & 40.383 & 33.047                  & 26.991                 & 35.416                    & 28.066                                       \\
~                        & \ding{51}                      & 25.542  & 31.544 & 37.040 & 40.688 & 33.261                  & 28.068                 & 36.183                    & 28.886~                                      \\
\ding{51}                       & \ding{51}                      & 25.530  & 31.522 & 37.030 & 40.656 & 33.311                  & 28.130                 & 36.332                    & 28.917~                                      \\
\midrule
\end{tabular}
\label{tab3}
\end{table*}
\vspace{-1.5em}

\begin{table*}
\centering
\caption{A comparison of state-of-the-art methods on the SNU-FILM, UCF101, DAVIS, Vimeo90K and VFITex sets. Pre-trained models are used for evaluation. For QVI, the trained network on Vimeo90K provided by \cite{kalluri2020flavr} is used. ST-MFNET is retrained on the Vimeo90K set for fair comparison. The suggestions in \cite{smith2017don} are used to train ST-MFNet with a larger batch size without a degradation in performance. Best performing model is in bold, second best is underlined.}
\label{tab4}
\resizebox{\textwidth}{!}{
\begin{tabular}{ccccccccccc} 
\hline
\multirow{2}{*}{Model} & \multicolumn{4}{c}{SNU-FILM}                                          & \multirow{2}{*}{UCF101} & \multirow{2}{*}{DAVIS} & \multirow{2}{*}{Vimeo90K} & \multirow{2}{*}{VFITex} & \multirow{2}{*}{Runtime (s)} & \multicolumn{1}{l}{\multirow{2}{*}{Params (M)}}  \\ 
\cmidrule{2-5}
                       & Extreme         & Hard            & Medium          & Easy            &                         &                        &                           &                         &                              & \multicolumn{1}{l}{}                             \\ 
\midrule
QVI                    & 24.956          & 31.099~         & 36.569          & 39.810          & 32.904                  & 27.260                 & 35.097                    & 27.689                  & \uline{0.438}                & 29.23                                            \\
FLAVR                  & 25.175          & 30.852          & 36.335          & 40.385          & 33.385                  & 27.419                 & 36.304                    & 28.290                  & 1.392                        & 42.06                                            \\
ST-MFNET               & \textbf{25.713} & \uline{31.456}  & \uline{36.935}  & \textbf{40.669} & \uline{33.388}          & \textbf{28.302}        & \uline{36.507}            & \uline{28.912}          & 1.494                        & 21.03                                            \\
VFIT-B                 & 25.492          & 31.022          & 36.495          & 40.499          & \textbf{33.449}         & 28.047                 & \textbf{36.960}           & 28.642                  & 1.528                        & 29.08                                            \\
VFIT-S                 & 25.434          & 31.057          & 36.490          & 40.411          & 33.355                  & 27.882                 & 36.484                    & 28.688                  & 0.909                        & 7.54                                             \\
Ours                   & \uline{25.53}   & \textbf{31.522} & \textbf{37.030} & \uline{40.656}  & 33.311                  & \uline{28.131}         & 36.332                    & \textbf{28.917}         & \textbf{0.239}               & 27.59                                            \\
\hline
\end{tabular}}
\vspace{-1.0em}
\end{table*}

\begin{table}
\centering
\vspace{-0.8em}
\caption{The runtime and total memory usage of the proposed method compared to other state-of-the-art baselines. Out of Operational Memory (OOM) indicates that memory usage requirements exceeded the maximum GPU memory available.}
\resizebox{\columnwidth}{!}{
\begin{tabular}{cccccccc} 
\midrule
Frame Size                 & \begin{tabular}[c]{@{}c@{}}Performance\\Information\end{tabular} & \begin{tabular}[c]{@{}c@{}}EDENVFI\\(Ours)\end{tabular}   & VFIT-S & VFIT-B & ST-MFNET             & FLAVR & QVI    \\ 
\midrule
\multirow{2}{*}{1920x1080} & Runtime (s)                                                      & 0.763  & 2.058  & 3.396  & \multirow{2}{*}{OOM} & 3.151 & 0.940  \\
                           & Memory (MB)                                                      & 4867   & 9543   & 13849  &                      & 11615 & 4301   \\ 
\midrule
\multirow{2}{*}{1280x720}  & Runtime (s)                                                      & 0.239  & 0.909  & 1.528  & 1.494                & 1.392 & 0.438  \\
                           & Memory (MB)                                                      & 2739   & 4731   & 8363   & 9629                 & 5791  & 2535   \\ 
\midrule
\multirow{2}{*}{640x360}   & Runtime (s)                                                      & 0.0564 & 0.232  & 0.384  & 0.3831               & 0.356 & 0.130  \\
                           & Memory (MB)                                                      & 1513   & 1985   & 2991   & 3155                 & 2311  & 1463   \\
\midrule
\end{tabular}}
\vspace{-1.5em}
\label{tab5}
\end{table}
\vspace{0.5em}
\subsection{Quantitative Evaluations}

As shown in Table~\ref{tab4}, our proposed model EDENVFI performs favourably on most benchmark sets compared to state-of-the-art methods. Relative to VFIT-S, EDENVFI performs 0.229dB better on VFITex which is a dataset rich in texture, 0.249dB better on DAVIS, 0.245dB better on SNU-FILM easy, 0.540dB better on SNU-FILM medium, 0.465dB better on SNU-FILM hard and 0.096dB better on SNU-FILM extreme. On Vimeo90K and UCF101, VFIT-S performs 0.152dB and 0.044dB better respectively. 

The results indicate that the proposed EDENVFI model performs better on scenes with dynamic motion and texture rather than the constrained motion observed in Vimeo90K. Although our model uses more parameters than VFIT-S, the main bottlenecks in practical implementations, runtime and memory usage are significantly reduced. Compared to other models such as FLAVR and ST-MFNET, our model generally performs well on datasets with more texture and dynamic motion. Whilst ST-MFNet achieves good performance on the test sets, the model uses significantly more memory.
\vspace{-0.6em}

\subsection{Runtime and Memory Usage}
For all frames of different spatial resolutions, our network EDENVFI has a quicker runtime compared to VFIT-S, VFIT-B, ST-MFNET, FLAVR and QVI, as evidenced in Table~\ref{tab5}. Experiments are conducted on the NVIDIA Quadro RTX 5000 with an Intel Xeon 5218R CPU. 

For 1920x1080 frames, a speedup by a factor of 2.70 is observed if our method is used instead of VFIT-S for frame synthesis. Our method is more than 1.2 times faster than QVI and more than 4 times faster than FLAVR, respectively. ST-MFNET requirements exceeded available GPU memory.
For smaller frame sizes, our model observes larger performance speedups relative to VFIT-S with our model being more than 4 times faster on 640x360 frames. Compared to ST-MFNET, our model is more than 6.5 times faster when synthesising a 640x360 frame.

In terms of memory usage on a 1920x1080 frame, our model uses 49$\%$ less memory than VFIT-S, almost 65$\%$ less memory compared to VFIT-B 58$\%$ less memory compared to FLAVR. For a 1280x720 frame, ST-MFNET uses approximately 9.6GB of GPU memory compared to the 2.7GB our model uses, more than a 70$\%$ reduction in memory usage.
\vspace{-0.7em}
\section{Conclusion}
\vspace{-0.5em}
This work presents an approach which integrates transformers into the video frame interpolation pipeline without significantly increasingcomputational burden. Our model performs favourably on a range of evaluation sets and is robust to dynamic motions observed in difficult sequences. The integration of PVT and the convolutional encoder enables the network to better handle different motions. Possible future improvements would involve finding a more effective way of fusing transformer and convolutional features. This could lead to further memory reductions.

\vfill\pagebreak

\bibliographystyle{IEEEbib}
\bibliography{egbib}

\end{document}